\documentclass[runningheads]{llncs}
\usepackage{graphicx}
\usepackage[colorlinks=true, citecolor=black, linkcolor=black]{hyperref} 

\usepackage[caption=false]{subfig}  % For captioning multi-panel figures
\usepackage{xcolor} 

\usepackage{longtable}
\usepackage{colortbl}
\usepackage{booktabs}
\usepackage{url}

\usepackage{amsmath}
\usepackage{tikz}
\usetikzlibrary{shapes,arrows,backgrounds,fit,positioning,calc,angles,quotes}

\usepackage{algorithm}
\usepackage{algpseudocode}
\usepackage{booktabs}
\usepackage{amssymb}



\begin{document}
\title{General Board Geometry}


\author{Cameron Browne \and {\'E}ric Piette \and Matthew Stephenson \and Dennis J. N. J. Soemers}
\authorrunning{C. Browne {\it et al.}}
\institute{Department of Data Science and Knowledge Engineering, Maastricht University, Paul-Henri Spaaklaan 1, 6229 EN, Maastricht, the Netherlands
\email{\{cameron.browne,eric.piette,matthew.stephenson,dennis.soemers\}\\@maastrichtuniversity.nl}}
\maketitle              % typeset the header of the contribution
\begin{abstract}
Game boards are described in the Ludii general game system by their underlying graphs, based on tiling, shape and graph operators, with the automatic detection of important properties such as topological relationships between graph elements, directions and radial step sequences. 
This approach allows most conceivable game boards to be described simply and succinctly. 
%This paper outlines the basic mechanism with examples. 

\end{abstract}

%%%%%%%%%%%%%%%%%%%%%%%%%%%%%%%%%%%%%%%%%%%%%%%%%%%%%%%%%%%%%

\section{Introduction}

The Digital Ludeme Project (DLP) is a five-year research project using Artificial Intelligence techniques to improve our understanding of the development of games throughout history \cite{Browne_2018_Modern}. 
We are modelling the 1,000 most ``important'' traditional strategy games in a consistent digital format, to provide a playable database of the world's traditional games for comparative analysis. 

The Ludii general game system\footnote{Ludii is available at \url{ludii.games} and the source code at \url{github.com/Ludeme/Ludii}} \cite{Piette_2020_Ludii} is a software tool developed specifically for this task, for modelling the full range of possible board games (950+ games implemented in version 1.2.8). 
Games are described in terms of simple {\it ludemes} assembled into structures to define arbitrarily complex behaviour, where each ludeme is a game-related concept implemented as a Java class (or enum attribute) in the Ludii code base~\cite{Browne_2016_Class}.

A key challenge in this task is to allow the user to describe arbitrarily complex game boards in a simple and intuitive way. 
This paper outlines our method for describing game boards in the Ludii grammar for general games.

%%%%%%%%%%%%%%%%%%%%%%%%%%%%%%%%%%%%%%%%%%%

\section{Game Graphs}

In Ludii, the board shared by all players is represented internally as a finite graph defined by a triple of sets $\mathcal{G} = \langle V, E, C \rangle$ in which $V$ is a set of {\it vertices}, $E$ a set of {\it edges}, and $C$ a set of {\it cells}. In graph theory, a cell is more commonly called {\it face} and represents a region bounded by a set of edges and that contains no other vertex or edge.\footnote{We sometimes use ``game design'' terms or definitions in lieu of stricter mathematical equivalents, in keeping with Ludii's primary purpose as a game design tool.} 
Vertex, edge and cell are all graph elements which can refer to each other, and denote {\it playable sites} at which players can place components during the game:
\begin{itemize}
    \item Let $v \in V$ denote a vertex. Then $v$ is an endpoint to each edge in $E(v)$, $C(v)$ gives the set of cells that $v$ is part of, and $V(v) = \{v\}$. %\varnothing$.
    \item Let $e \in E$ denote an edge. Then $V(e)$ is a set of 2 vertices that are the endpoints of $e$, $C(e)$ gives the set of cells $e$ is bounding, and $E(e) = \{e\}$. %\varnothing$.
    \item Let $c \in C$ denote a cell. Then $E(c)$ is the set of all the edges bounding $c$, $V(c)$ gives the set of the vertices which are the endpoints of the edges bounding $c$, and $C(c) = \{c\}$. % \varnothing$.
\end{itemize}

\vspace{-4mm}

\begin{figure}[h!tbp]
\centering
%\begin{minipage}[b]{0.32\textwidth} 
%    \subfloat[\label{fig:GraphElementA}]{%
%        \includegraphics[width=.95\textwidth]{figs/CellEdgeVertex-1.PNG}
%    }
%    % \includegraphics[width=\textwidth]{figs/CellEdgeVertex.PNG}
%    % \subcaption{} 
%    % \label{fig:GraphElementA}
%\end{minipage}
%\hspace{0.5cm}
\begin{minipage}[b]{0.36\textwidth} 
    \subfloat[\label{fig:GraphElementB}]{%
        \includegraphics[width=.95\textwidth]{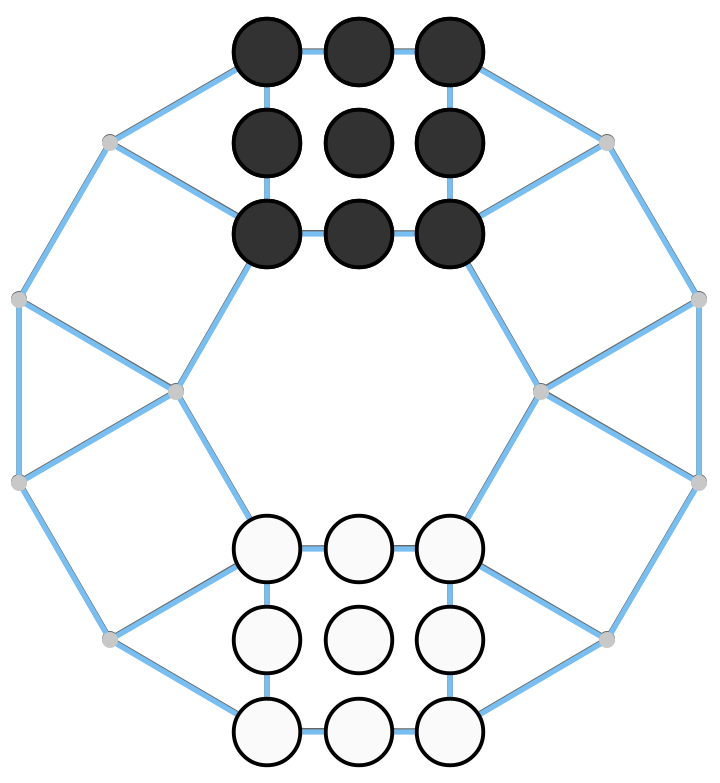}
    }
% 	\includegraphics[width=\textwidth]{figs/Triple Tangle.PNG}
%     \subcaption{} 
%     \label{fig:GraphElementB}
\end{minipage}
\hspace{1cm}
\begin{minipage}[b]{0.45\textwidth} 
    \subfloat[\label{fig:GraphElementC}]{%
        \includegraphics[width=.95\textwidth]{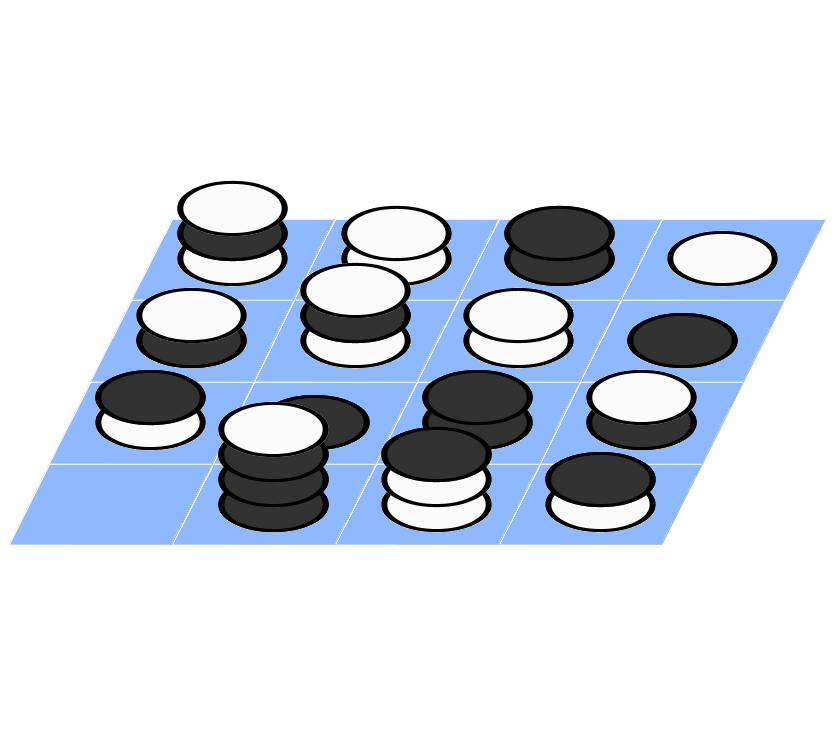}
    }
% 	\includegraphics[width=\textwidth]{figs/Score Four.PNG}
%     \subcaption{} 
%     \label{fig:GraphElementC}
\end{minipage}
\caption{A game played on vertices, edges and cells (a) and a game played on cells (b).}
\label{fig:GraphElement}
\end{figure}

%Figure \ref{fig:GraphElementA} shows each of the three graph element types on a graph modelling a single square tiling. 
For example, Figure \ref{fig:GraphElementB} shows a game with pieces played on the vertices, edges and cells of the board graph. 
Figure \ref{fig:GraphElementC} shows a board game played only on the cells but in which pieces may stack.

In any single game, components (or a stack of components) can be placed on any graph element. For this reason, we define a playable site as a triple $\langle Type, Index, Level \rangle$ in which the $Type$ can be (Vertex, Edge or Cell), the $Index$ is the number of the element and the $Level$ is the index of the element in the stack ($0$ meaning the ground).

Any ludeme referring to a playable site has to specify each of these data. However, for convenience, Ludii uses default values. The default type of a location is Cell, except if the description of the game specifies another default site type. For levels, the default value is the top Level of the location specified, as stacked site are typically owned by the player with a piece on top.

\subsection{Dimensions: Cells or Vertices}

\begin{figure}[h!tbp]
\centering
\begin{minipage}[b]{0.45\textwidth} 
    \subfloat[\label{fig:Chess}]{%
        \includegraphics[width=1\textwidth]{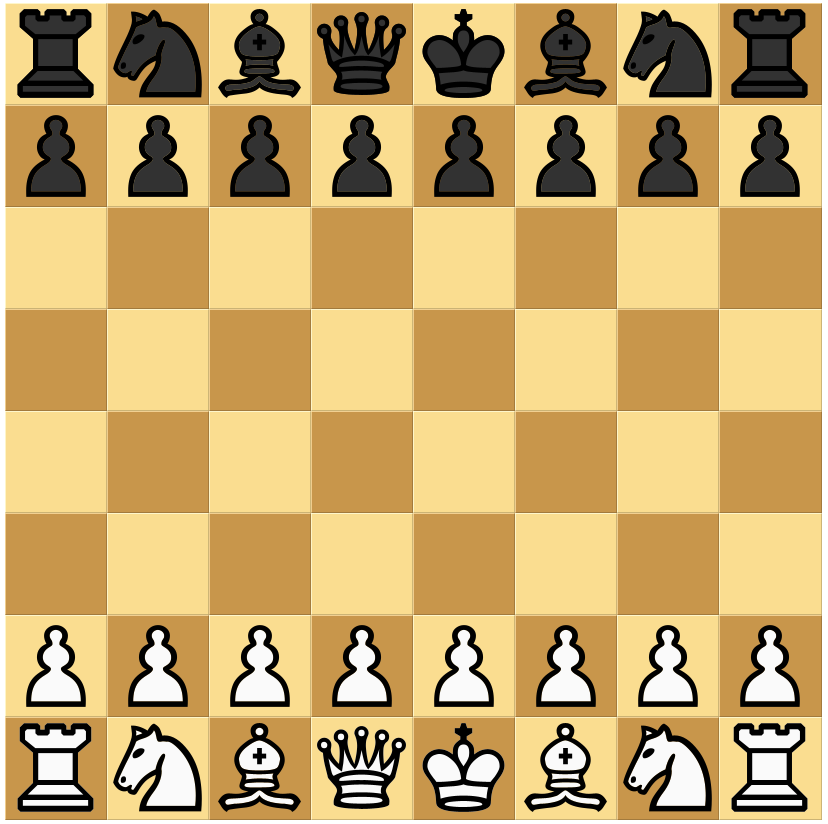}
    }
    % \includegraphics[width=\textwidth]{figs/Chess.PNG}
    % \subcaption{} 
    % \label{fig:Chess}
\end{minipage}
\hspace{0.5cm}
\begin{minipage}[b]{0.45\textwidth} 
    \subfloat[\label{fig:Go}]{%
        \includegraphics[width=1.025\textwidth]{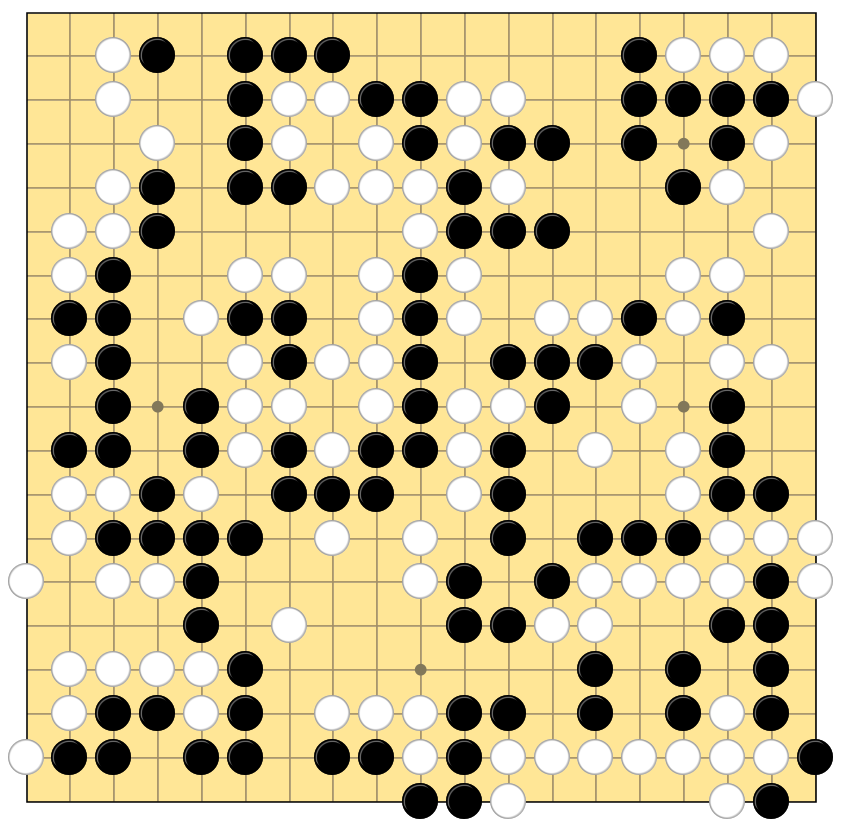}
    }
% 	\includegraphics[width=\textwidth]{figs/Go.PNG}
%     \subcaption{} 
%     \label{fig:Go}
\end{minipage}
\caption{(a) Chess (8x8 Cells). (b) Go (19x19 Vertices).}
\label{fig:ChessGo}
\end{figure}

The graph is generated based on the specified board dimensions and default site type. For example, a Chess board described {\tt (board (square 8))} (see Figure \ref{fig:Chess}) produces a square grid with $8$ cells per row and column. However, a Go board described as {\tt (board (square 19) use:Vertex)} (see Figure \ref{fig:Go}) produces a square grid with $19$ vertices per row and column. If the default site type is {\tt Vertex} or {\tt Edge} then the board dimensions are based on the number of vertices rather than cells.

\vspace{-5mm}

%%%%%%%%%%%%%%%%%%%%%%%%%%%%%%%%%%%%%%%%%%%

\section{Game Board Description}

Game boards are described in the Ludii grammar \cite{Browne_2020_LLR} using the following basic EBNF syntax: {\tt <board> ::= (board <graph>)} where the underlying {\tt <graph>} object defines the vertices, edges and cells that make up the game board. 

\begin{figure}[h!tbp]
\centering
\begin{minipage}[b]{0.32\textwidth} 
    \subfloat[\label{fig:BoardsA}]{%
        \includegraphics[width=1\textwidth]{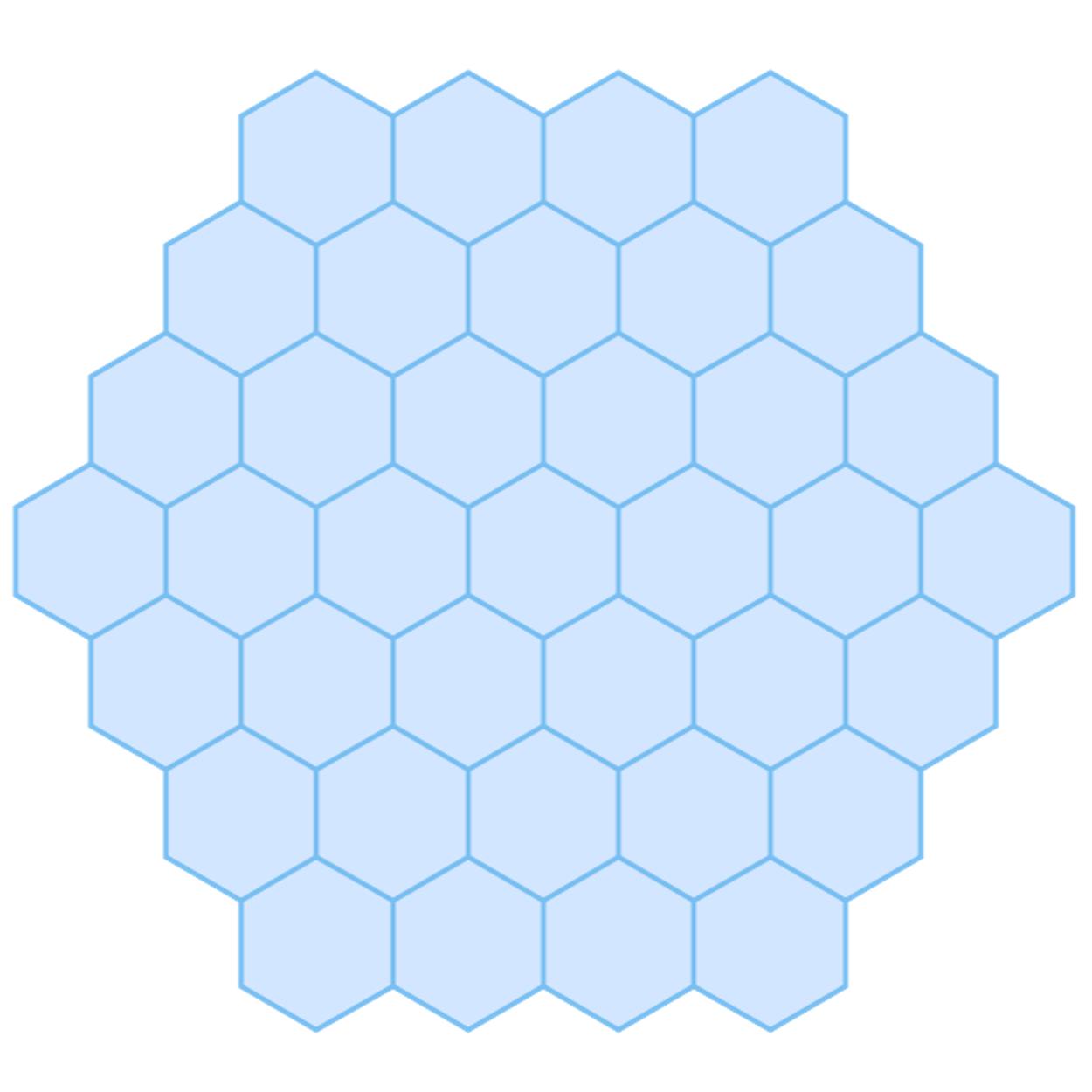}
    }
% 	\includegraphics[width=\textwidth]{figs/tiling-6.png}
%         	\subcaption{} 
%         	\label{fig:BoardsA}
\end{minipage}
%\hspace{0.5cm}
\begin{minipage}[b]{0.32\textwidth} 
    \subfloat[\label{fig:BoardsB}]{%
        \includegraphics[width=1\textwidth]{figs/tiling-3464.PNG}
    }
% 	\includegraphics[width=\textwidth]{figs/tiling-3464.png}
%         	\subcaption{} 
%         	\label{fig:BoardsB}
\end{minipage}
%\hspace{0.5cm}
\begin{minipage}[b]{0.32\textwidth} 
    \subfloat[\label{fig:BoardsC}]{%
        \includegraphics[width=1\textwidth]{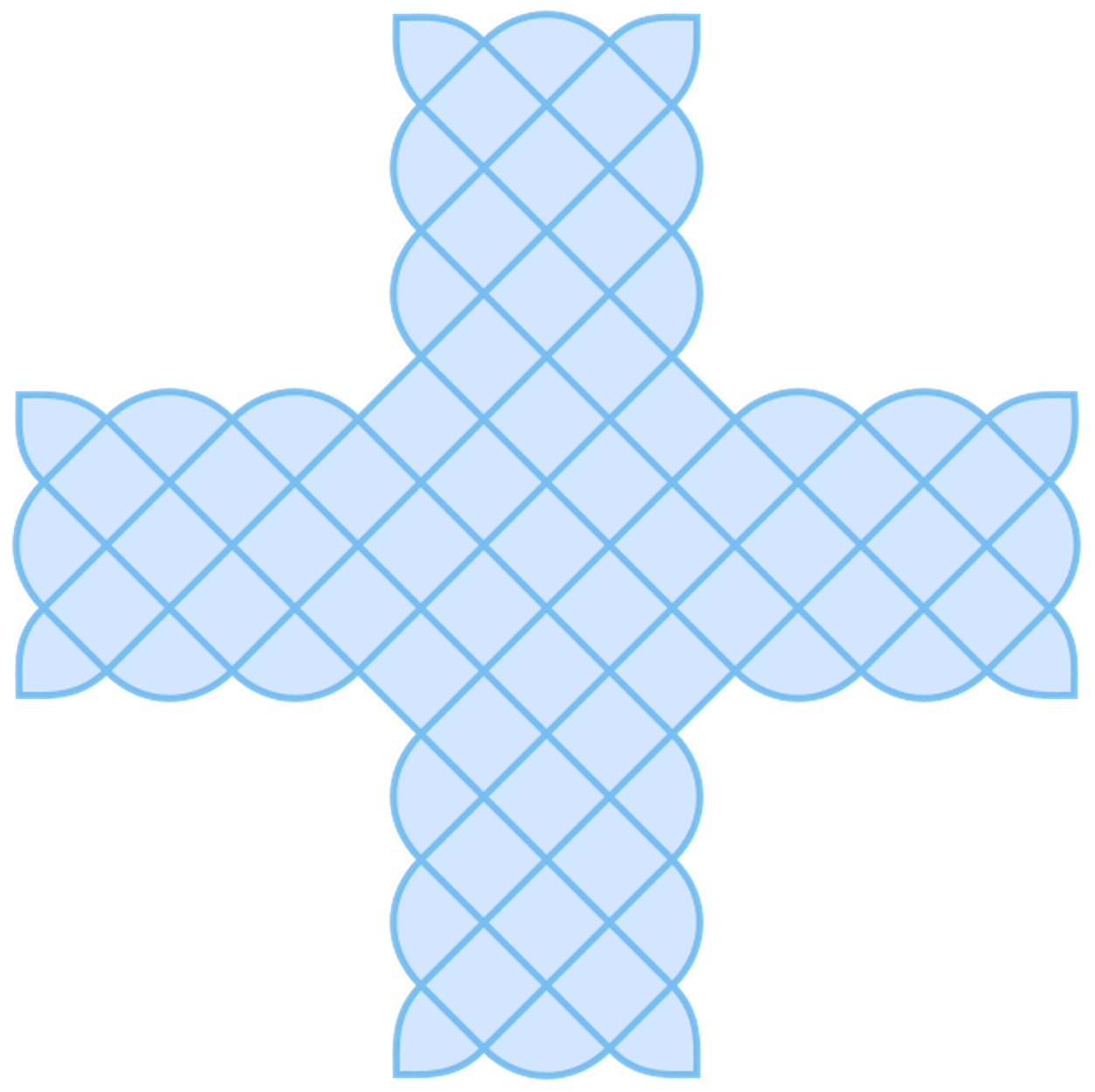}
    }
% 	\includegraphics[width=\textwidth]{figs/celtic-1.png}
%         	\subcaption{} 
%         	\label{fig:BoardsC}
\end{minipage}
\caption{Boards from tilings: hexagonal (a), semi-regular 3.4.6.4 (b) and {\tt celtic} (c).}
\label{fig:Boards}
\end{figure}

The user can specify the location of each vertex (and adjacencies between them as edges) to allow the description of arbitrarily complex graphs, or they can take advantage of a range of predefined tilings, shapes and graph operators for more concise descriptions (described more fully in Section~\ref{sec:GraphOperators}). 
For example, the three game boards shown in Figure \ref{fig:Boards} are described by the following graphs (the {\tt poly} field describes the polygonal shape of the board):

\phantom{}

\noindent
{\tt (hex 4)}

\phantom{}

\noindent
{\tt (tiling T3464 2)}

\phantom{}

\noindent
{\tt (celtic (poly \{\{3 0\}\{3 4\}\{0 4\}\{0 7\}\{3 7\}\{3 11\}\{6 11\} }

{\tt \hspace{2.05cm} \{6 7\}\{10 7\} \{10 5\} \{6 5\}\{6 0\}\}))}

%%%%%%%%%%%%%%%%%%%%%%%%%%%%%%%%%%%%%%%%%%%

\section{Graph Relations}

For a graph $\mathcal{G} = \langle V, E, C \rangle$, two different graph elements $g_1$ and $g_2$ can have different relations:
\begin{itemize}

    \item \textbf{Adjacent}: $g_1$ and $g_2$ are {\it adjacent} if and only if $\left( \exists e \in E(g_1) \cap E(g_2) \right) \lor \left( \exists v \in V(g_1) \cap V(g_2) \right) \lor \left( \exists c \in C(g_1) \cap C(g_2) \right)$. In other words, two graph elements are adjacent if they share any graph element they are referring.
    
    \item \textbf{Orthogonal}: $g_1$ and $g_2$ are {\it orthogonal} if and only if $\exists e_1 \in E(g_1), \exists e_2 \in E(g_2), e_1 = e_2$. In other words, two graph elements are orthogonal if they share an edge.
    
    \begin{figure}[h!tbp]
\centering
\begin{center}
\includegraphics[scale=1.0]{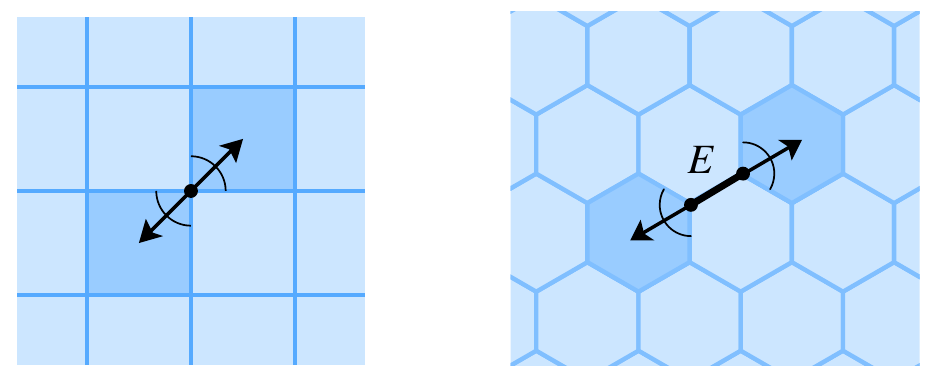}
\end{center}
\caption{Adjacent diagonals (left) and non-adjacent diagonals (right).}
\label{fig:Diagonals}
\end{figure}

\vspace{-2mm} 

    \item \textbf{Diagonal}: Two cells are considered {\it diagonal} if and only if:\footnote{This definition differs slightly from the actual implementation, but it captures the general understanding of {\it diagonality} between cells on a game board.}
    \begin{enumerate}
        \item They share a vertex (but not an edge) and the bisectors of the angles at that vertex in each cell are maximally opposed. Note that a cell can have multiple diagonal neighbours through a vertex if all satisfy this property. Or:
        \item If a cell has no such adjacent neighbour through a given vertex, then we allow a non-adjacent diagonal neighbour through that vertex if the two cells are coincident with the end points of some edge $E$ (which does not belong to either cell) and the bisectors of the angles at the end point in each cell are maximally opposed. 
    \end{enumerate}
    These two diagonal relationships are shown in Figure~\ref{fig:Diagonals}.
    
    Diagonality is defined similarly for vertices, but transposing ``cell'' and ``vertex'' in the above definitions.

    \item \textbf{Off Diagonal}: $g_1$ and $g_2$ are off diagonal if and only if $g_1 \in C, g_2 \in C,\exists v_1 \in V(g_1), \exists v_2 \in V(g_2), v_1 = v_2,\nexists e_1 \in E(g_1), \nexists e_2 \in E(g_2), e_1 = e_2$. In other words, two cells are off diagonal if they are not diagonal, not orthogonal and they share a vertex.
    \item \textbf{All}: $g_1$ and $g_2$ are related if they are orthogonally, adjacently, diagonally or off diagonally related to each other.
\end{itemize}

These relationships are summarised for the regular tilings in Table~\ref{tab:GraphRelations}.
%Figure~\ref{fig:Diagonals} shows the two types of diagonal relationships. 

\begin{table}[h!tbp]
\caption{Relations for the regular tilings.}
\centering
\begin{tabular} { m{20mm} m{105mm} }
\toprule
{\bf Relation} & \hspace{7.5mm} {\bf Square} \hspace{21mm} {\bf Triangular} \hspace{18mm} {\bf Hexagonal}  \\
\midrule
All          & \includegraphics[width=1\linewidth]{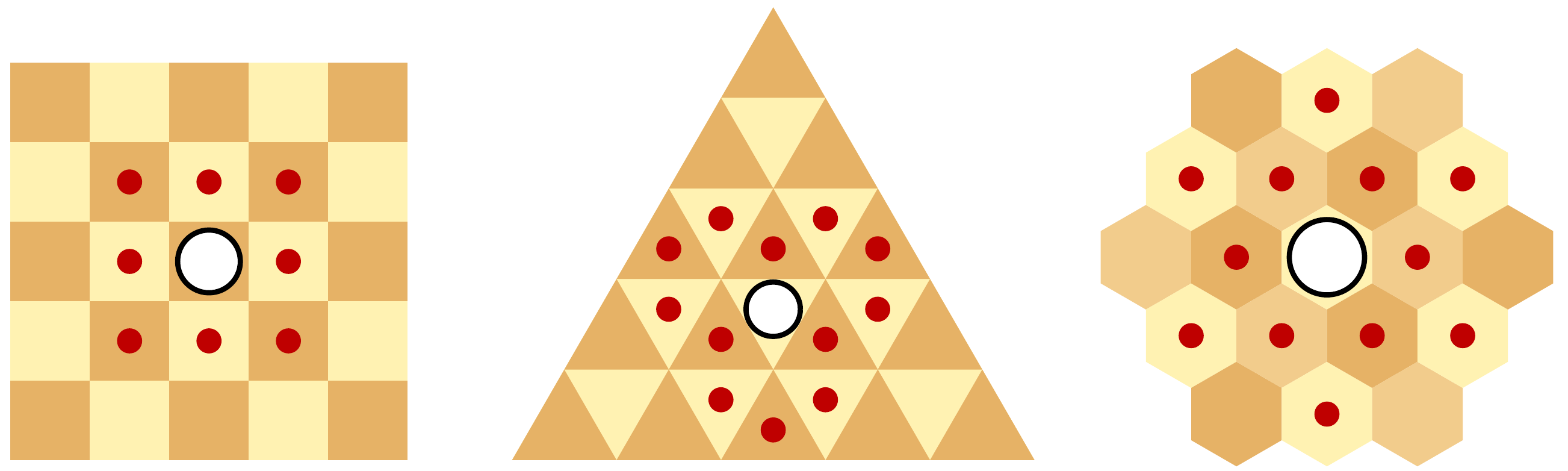} \\
Adjacent     & \includegraphics[width=1\linewidth]{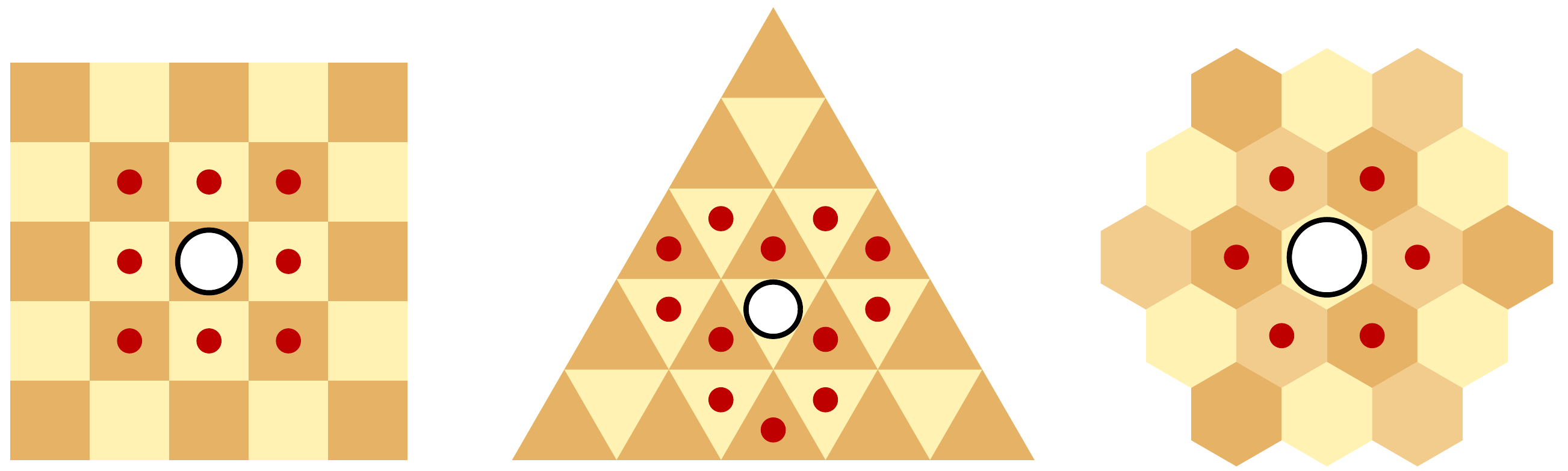} \\
Orthogonal   & \includegraphics[width=1\linewidth]{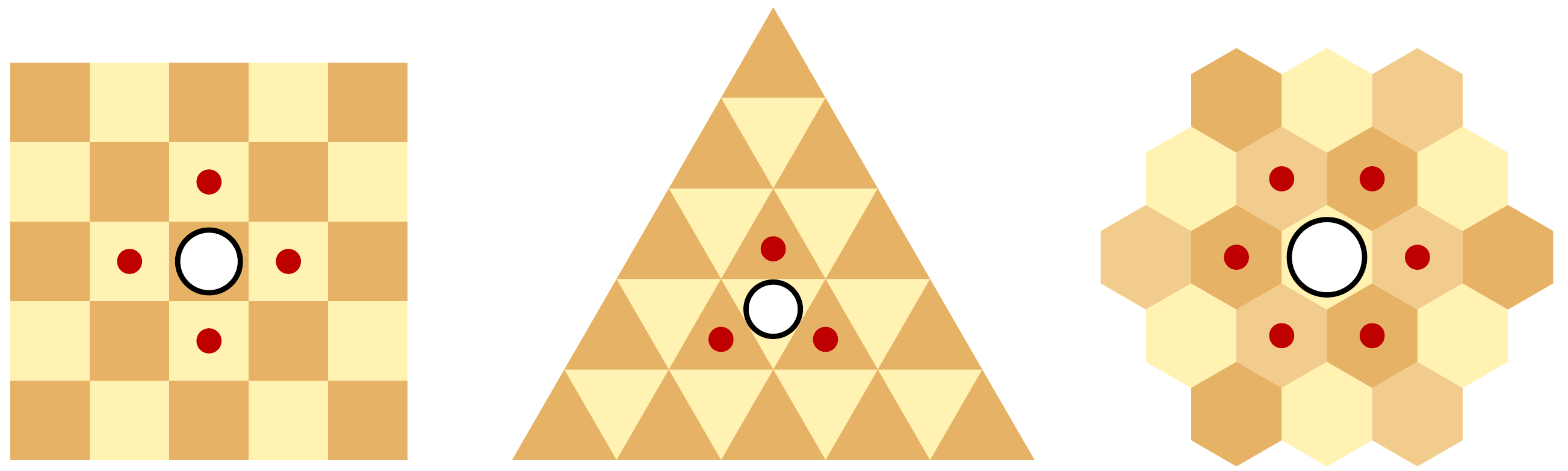} \\
Diagonal     & \includegraphics[width=1\linewidth]{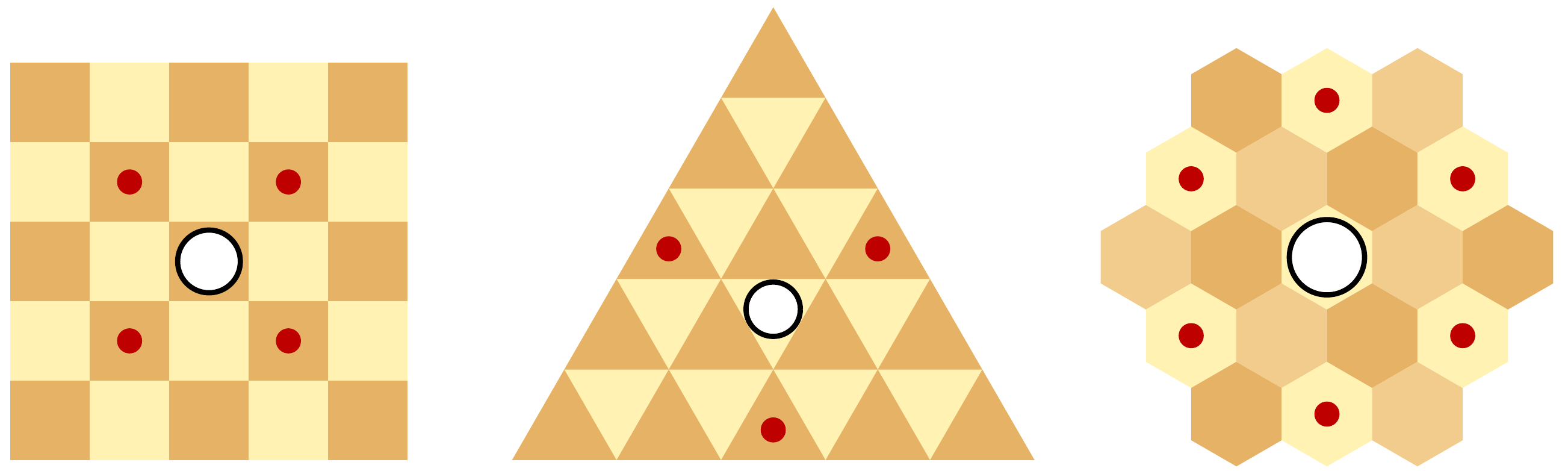} \\
Off-Diagonal & \includegraphics[width=1\linewidth]{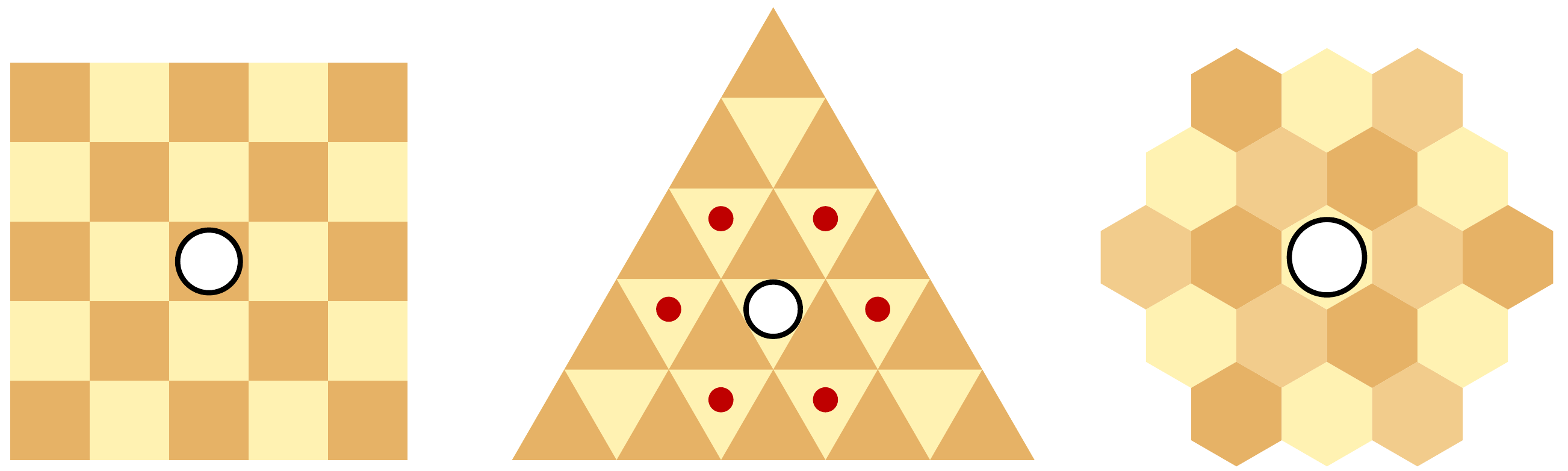} \\
\bottomrule
\end{tabular}
\label{tab:GraphRelations}
\end{table}

%----------------------------

\subsection{Directions}

%\begin{figure}[h!tbp]
%\centering
%\begin{center}
%\includegraphics[scale=0.15]{figs/intercardinal.png}
%\end{center}
%\caption{Intercardinal directions: %\href{https://en.wikipedia.org/wiki/Cardinal_direction}{Wikipedia}}
%\label{fig:IntercardinalDirections}
%\end{figure}

Ludii supports the following direction types:

\begin{itemize}

    \item {\it Intercardinal} directions: {\tt N, NNE, NE, ENE, E, ESE, SE, SSE, S, SSW, \\ SW, WSW, W, WNW, NW, NNW.} 
    
    \item {\it Rotational} directions: {\tt In, Out, CW} (clockwise), {\tt CCW} (counter-clockwise). 
    
    \item {\it Spatial} directions for 3D games: {\tt D, DN, DNE, DE, DSE, DS, DSW, DW, \\ DNW and U, UN, UNE, UE, USE, US, USW, UW, UNW.} 
    
    \item {\it Axial} directions subset (for convenience): {\tt N, E, S, W.}
    
    \item {\it Angled} directions subset (for convenience): {\tt NE, SE, SW, NW.}

\end{itemize}

%Moreover, a bunch of other directions exist such as {\it Axial} corresponding to $\{N, S, E, W\}$ or {\it Angled} corresponding to $\{NE, NW, SE, SW\}$.

%up to 16 compass directions following the intercardinal directions (Figure \ref{fig:IntercardinalDirections}) and the 4 rotational directions: Outer (Out), Inner (In), Clockwise (CW) and Counter Clockwise (CCW). Concerning the 3D games, Ludii handles many spatial directions which are included all the compass directions for the same layer and the 8 most important compass directions for the downwards and upwards directions: D, DN, DNE, DE, DSE, DS, DSW, DW, DNW and U, UN, UNE, UE, USE, US, USW, UW, UNW.

Each graph element $g$ has a corresponding set of {\it absolute directions} $A_d$ and {\it relative directions} $R_d$ to associated graph elements of the same type. 
Absolute directions can be any of the above direction types in addition to any relation type ({\tt Adjacent, Orthogonal, Diagonal, Off Diagonal}, or {\tt All}). 

Relative directions from an element $g$ are defined by  $R_d(g, facing, rotation,$ $relation)$ where $facing$ describes the direction in which a component at $g$ is facing, $rotation$ describes the number of rightward steps of the component at $g$, and $relation$ describes the graph relation to use at each step ({\tt Adjacent} by default). 
Relative directions are: {\tt Forward, Backward, Rightward, Leftward, FR, FRR, FRRR, FL, FLL, FLLL, BR, BRR, BRRR, BL, BLL} or {\tt BLLL}.

For example, consider a piece on a {\tt square} board (which involves only the eight major compass directions as adjacent relations). If the piece is facing {\tt N} (North) with a rotation of $0$, the relative direction {\tt Forward} is the graph element immediately to the North (upwards) if such an element exists. 
However, if that piece is facing {\tt E} (East) and its current rotation is $1$, the relative direction {\tt FR} (meaning ``Forward Right'') is the graph element to its South East (if such an element exists).

%----------------------------

\subsection{Steps and Walks}

A {\it step} is a record of two related graph elements ({\it from} and {\it to}) which can be of different types and the absolute directions that describe their relationship. 
For example, a cell $A$ directly above another cell $B$ on a Chess board could be described as an {\tt Adjacent}, {\tt Orthogonal} or {\tt N} step away.

\begin{figure}
\centering
\includegraphics[width=\textwidth]{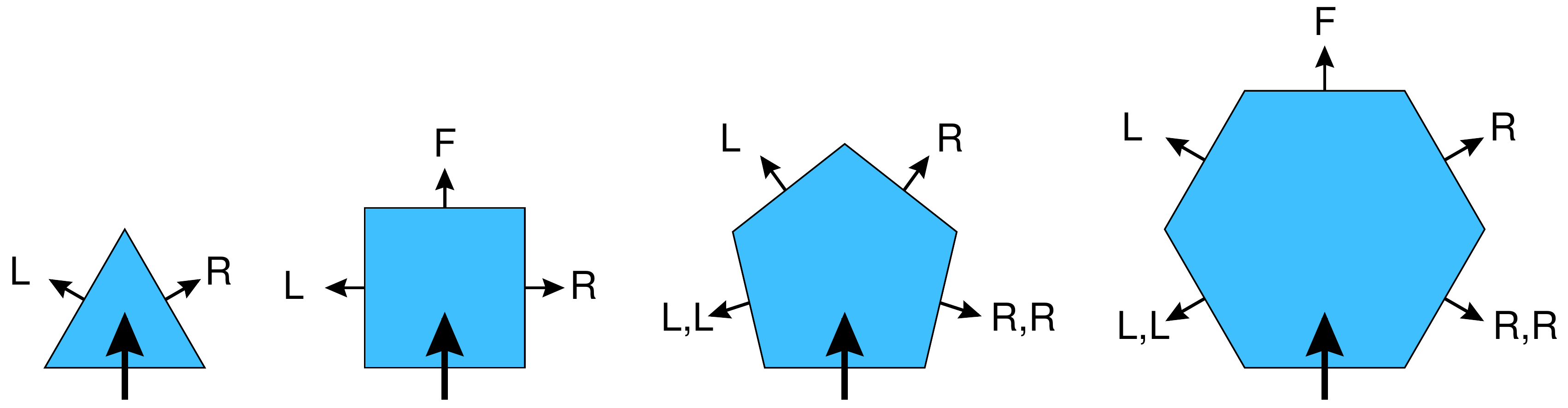}
\caption{Relative steps from various cell types.}
\label{fig:Steps}
\end{figure}

Ludii also provides three relative step types ({\tt F}, {\tt L} and {\tt R}) that allow users to define {\it walks} within the board graph. 
These correspond the standard ``forward'', ``left'' and ``right'' commands used in {\it turtle graphics}~\cite{Abelson_1986_Turtle}, as shown in Figure~\ref{fig:Steps}.

\vspace{-2mm}

\vspace{-2mm}

\begin{figure}[h!]
\centering
\includegraphics[width=0.65\textwidth]{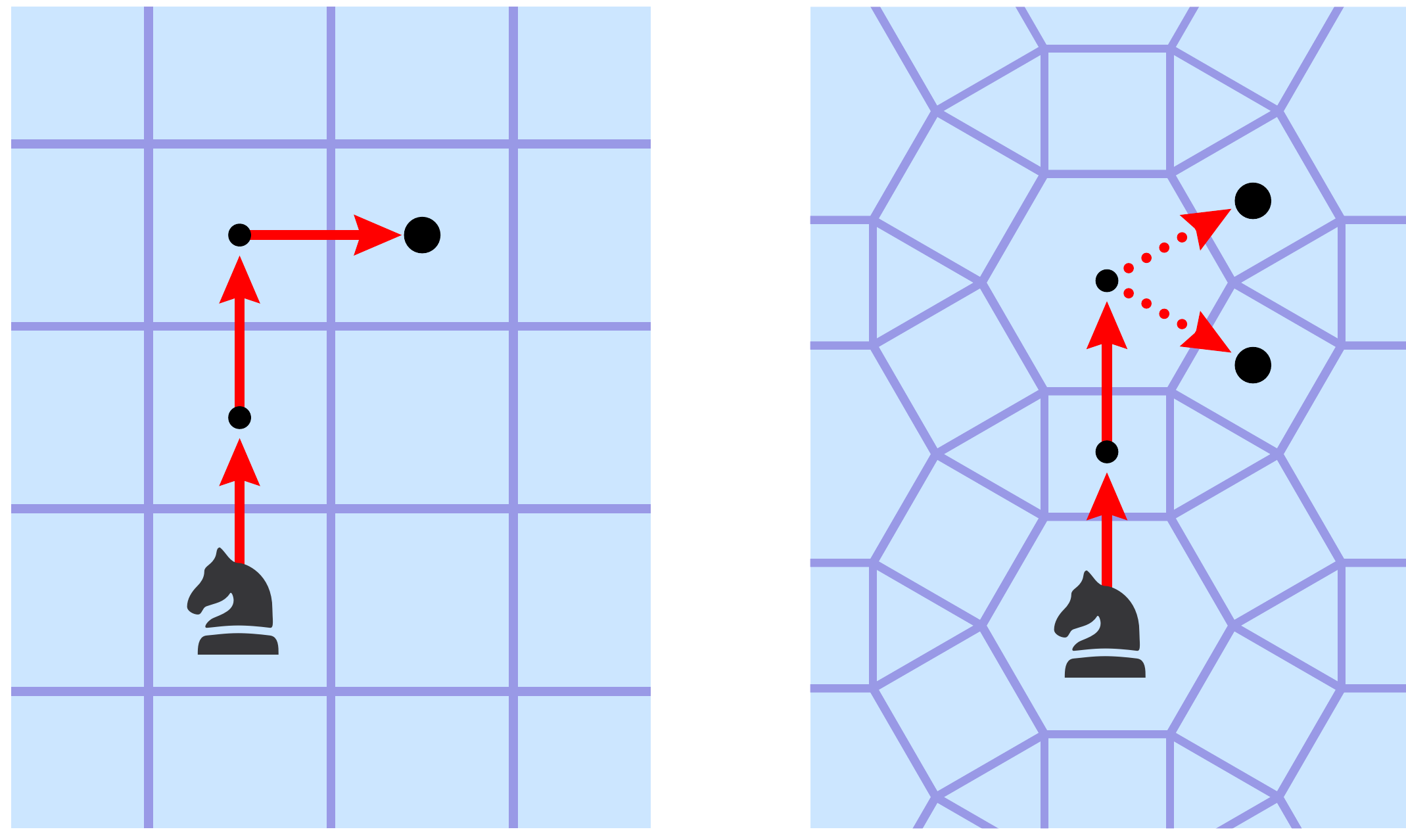}
\caption{Walk {\tt \{F,F,R\}} describes knight moves on square and 3.4.6.4 tilings.}
\label{fig:Walk}
\end{figure}

This representation allows descriptions of piece movements to be easily transferred between different board topologies. 
For example, a knight move in Chess may be described as the walk {\tt \{F,F,R\}} as shown in Figure~\ref{fig:Walk} (left) ans this walk may be directly used on a board based on the semi-regular 3.4.6.4 tiling (Figure~\ref{fig:Walk}, right). 
Note, however, that different topologies may introduce ambiguities such as whether both right turns in the 3.4.6.4 knight move (dotted lines) should be considered valid moves or only one of them (probably the furthest reaching one). 
Such ambiguities should be resolved by the game designer according to the behaviour they want.

\vspace{-1mm}

%----------------------------

\subsection{Radials}

Many games involve piece movement through contiguous lines of cells in a direction, such as slide moves by the queen, rook and bishop pieces in Chess. Such lines of play are called {\it radials}.  
Ludii automatically pre-generates all possible steps between playable sites on the board and all possible radials derived from them, for convenient game description and efficient processing.

\begin{figure}
\centering
\includegraphics[width=0.55\textwidth]{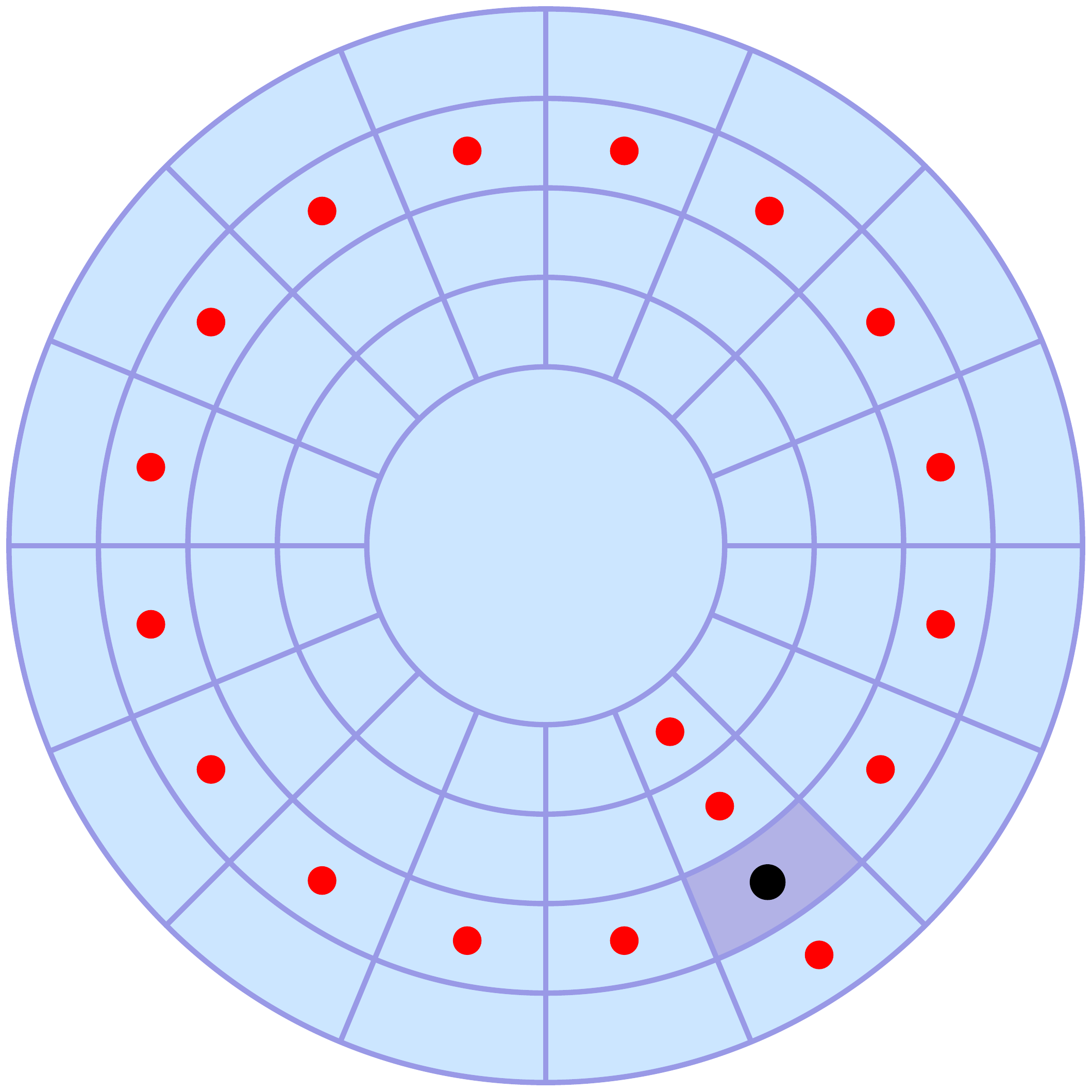}
\caption{Orthogonal radials on a circular Chess board (Shatranj ar-Rumiya).}
\label{fig:CircularRadials}
\end{figure}

For each playable site on the board $S$, each valid step to a neighbouring graph element of the same type in an absolute direction $d$ is extended as far as possible, to produce a radial from $S$ in direction $d$. 
For example, Figure~\ref{fig:CircularRadials} shows {\tt Orthogonal} radials from the shaded cell on a circular Chess board, such as a rook would move in the game Shatranj ar-Rumiya. 
Note that radials may bend to follow the board topology.

Radials extend step-by-step in the given absolute direction that minimises deviation in the radial's current heading. 
If the next step would deviate by 90$^{o}$ or more, then the radial terminates.

\begin{figure}
\centering
\includegraphics[width=0.55\textwidth]{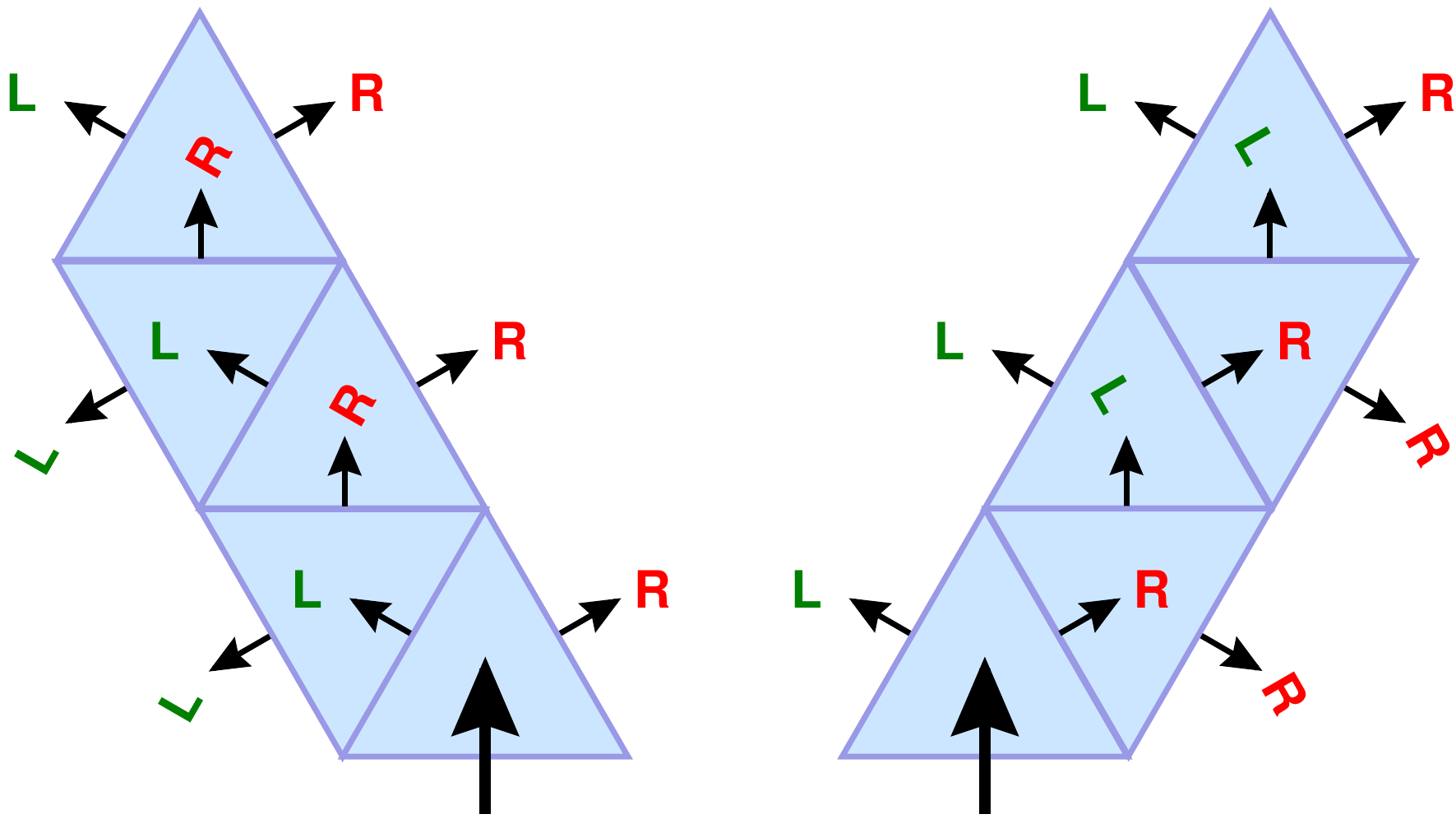}
\caption{Branching radial on a triangular grid.}
\label{fig:TriangularRadials}
\end{figure}

Radials can {\it branch} where two or more steps in the current direction are equally as good.\footnote{Still to be implemented in Ludii.}  
For example, Figure~\ref{fig:TriangularRadials} shows how an {\tt Orthogonal} step into a triangular cell may validly continue either {\tt L} (left) or {\tt R} (right), and thereafter alternate {\tt \{L,R,L,R,...\}} to produce branching zig-zagging radials in which the direction of each individual step is less important than the average direction of the radial overall. 
 
\vspace{-2mm}

%%%%%%%%%%%%%%%%%%%%%%%%%%%%%%%%%%%%%%%%%%%

\section{Graph Operators}  \label{sec:GraphOperators}

\noindent
Graphs are initially defined by a tiling and/or shape but can then be further modified using a range of {\it graph operators}. 
The complete set of tilings, shapes and graph operators defined in the Ludii grammar is shown in Table \ref{tab:Keywords}.
These can be used in combination to define thousands of different boards types quickly and easily. 
Greyed out items indicate planned future work not implemented yet. 

%------------------------

\begin{table}[h]
\caption{Keywords in the Ludii grammar for describing game boards.}
\small
\begin{center}
\begin{tabular} { p{4cm} p{4cm} p{3cm} }
\toprule
\textbf{\bf Tiling} & {\bf Shape} & {\bf Operator} \\ 
\midrule

{\it Regular}             & {\tt square}         & {\tt add} \\   
{\tt square}              & {\tt rectangle}                     & {\tt clip} \\
{\tt hex}                   & {\tt hexagon}                         & {\tt complete} \\ 
{\tt tri}                      & {\tt triangle}                         & {\tt dual} \\
                               & {\tt wedge}          & {\tt hole} \\
{\it Semi-Regular}   &  {\tt regular} (polygon)  & {\tt intersect} \\
{\tt T488} ({\it i.e.} 4.8.8) & {\tt {\tt poly} (any polygon)}                          & {\tt keep} \\
{\tt T4612} ({\it i.e.} 4.6.12) &                      & {\tt layers} \\
{\tt T3464} ({\it i.e.} 3.4.6.4) &  {\it Attribute}         & {\tt makeFaces} \\
{\tt T3636} ({\it i.e.} 3.6.3.6) &  {\tt Star}          & {\tt merge} \\
{\tt T31212} ({\it i.e.} 3.12.12)  & {\tt Diamond} & {\tt recoordinate} \\
{\tt T33336} ({\it i.e.} 3.3.3.3.6) & {\tt Prism}                & {\tt remove} \\
{\tt T33344} ({\it i.e.} 3.3.3.4.4) &                 & {\tt renumber} \\
{\tt T33434} ({\it i.e.} 3.3.4.3.4) &  {\it Modifier}               & {\tt rotate} \\
                                & {\tt diagonals:<DiagType>}             & {\tt scale} \\
{\it Custom}             & {\tt pyramidal:<boolean>}   & {\tt shift} \\
{\tt concentric}         &  {\tt limping:<boolean>}     & {\tt skew} \\
{\tt spiral}                 &  \textcolor{lightgray}{{\tt fractal/recursive}}  & {\tt splitCrossings} \\
{\tt quadhex}           & \textcolor{lightgray}{{\tt lattice}}   & {\tt subdivide} \\
{\tt brick}                  & \textcolor{lightgray}{{\tt projective}}  & {\tt trim} \\
{\tt celtic}                 &   & {\tt union} \\
{\tt repeat}                 &   &  \\

\bottomrule
\end{tabular}
\label{tab:Keywords}
\end{center}
\end{table}

%------------------------

\begin{figure}[h!tbp]
\centering
\begin{minipage}[b]{0.32\textwidth} 
    \subfloat[\label{fig:DualA}]{%
        \includegraphics[width=.95\textwidth]{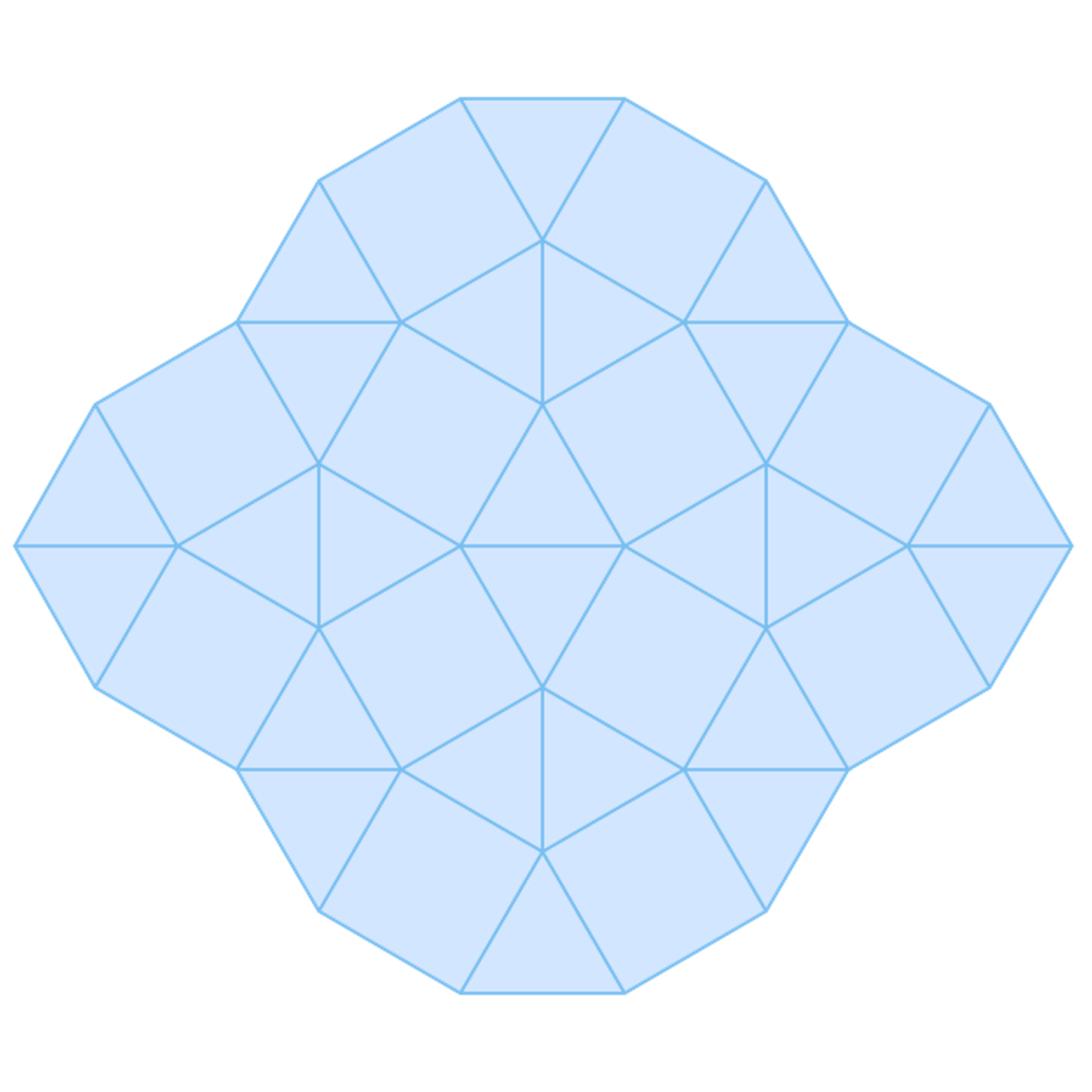}
    }
% 	\includegraphics[width=\textwidth]{figs/tiling-33434.png}
%         	\subcaption{} 
%         	\label{fig:DualA}
\end{minipage}
\hspace{0.1cm}
\begin{minipage}[b]{0.32\textwidth}
    \subfloat[\label{fig:DualB}]{%
        \includegraphics[width=.95\textwidth]{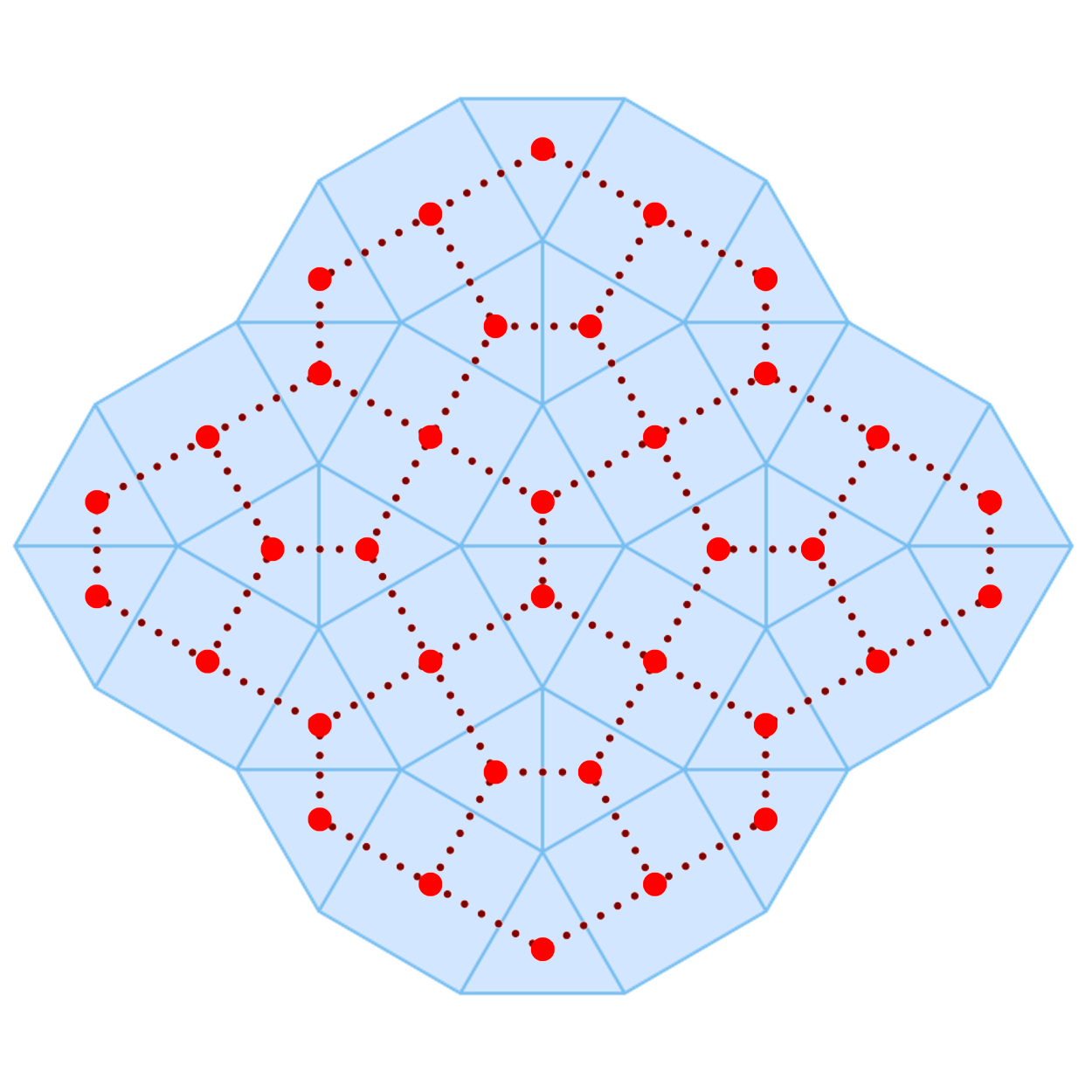}
    }
% 	\includegraphics[width=\textwidth]{figs/tiling-33434-dots.png}
%         	\subcaption{} 
%         	\label{fig:DualB}
\end{minipage}
%\hspace{0.2cm}
\begin{minipage}[b]{0.32\textwidth} 
    \subfloat[\label{fig:DualC}]{%
        \includegraphics[width=.95\textwidth]{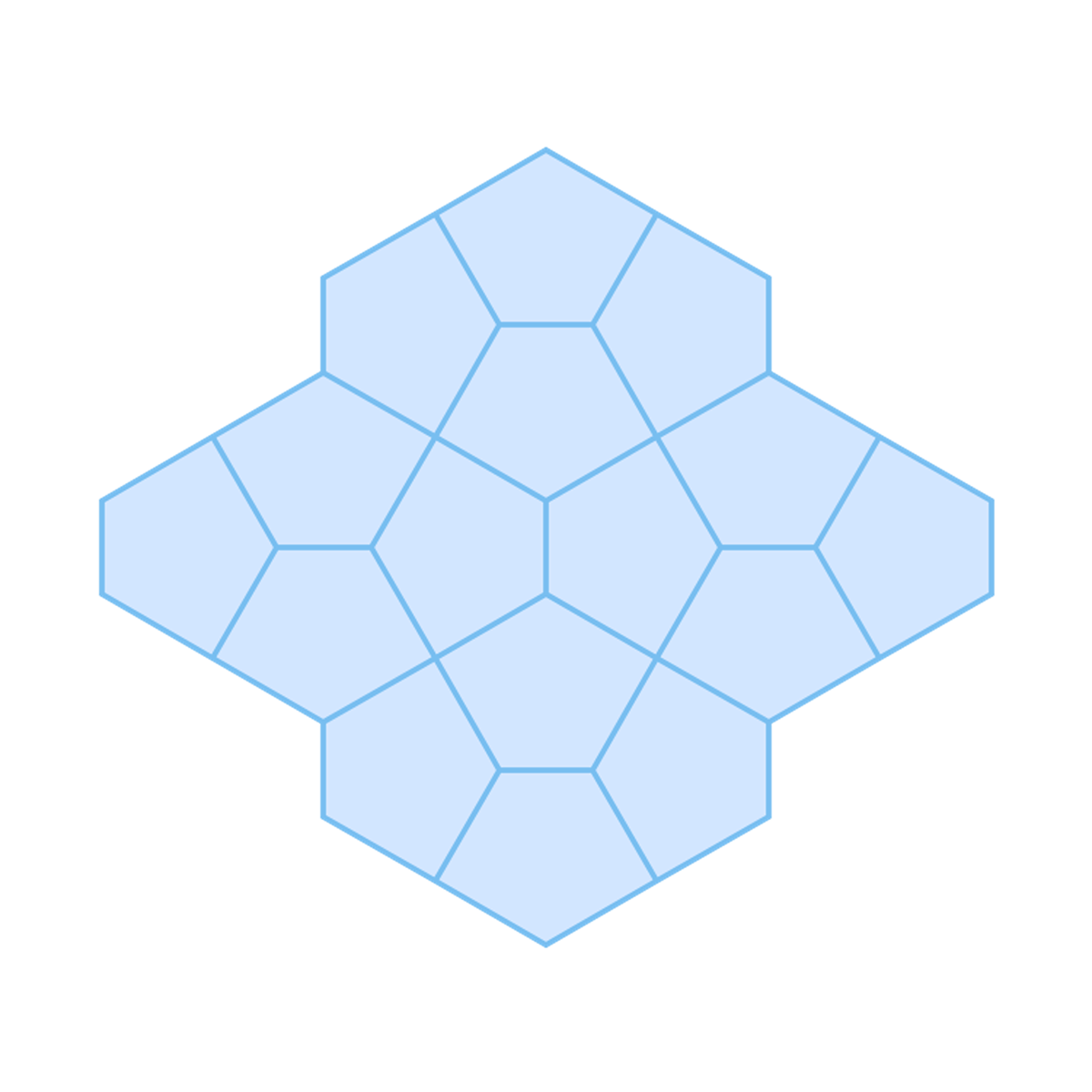}
    }
% 	\includegraphics[width=\textwidth]{figs/tiling-33434-dual.png}
%         	\subcaption{} 
%         	\label{fig:DualC}
\end{minipage}
\caption{A 3.3.4.3.4 tiling (a), its cell adjacencies (b) and its weak dual (c).}
\label{fig:Dual}
\end{figure}

For example, the very useful {\tt dual} operator converts a source graph into its {\it weak dual} defined by edges whose end points are the centroids of its adjacent cells. 
Figure~\ref{fig:Dual} shows a {\tt dual} operation applied to a small graph based on tiling 3.3.4.3.4 to produce the well known Cairo tiling:

\phantom{}

{\tt (dual (tiling T33434 2))}

%\phantom{}

\vspace{-4mm}

\vspace{-2mm}

\begin{figure}[h!]
\centering
\begin{minipage}[b]{0.32\textwidth} 
    \subfloat[\label{fig:SubdivideA}]{%
        \includegraphics[width=.95\textwidth]{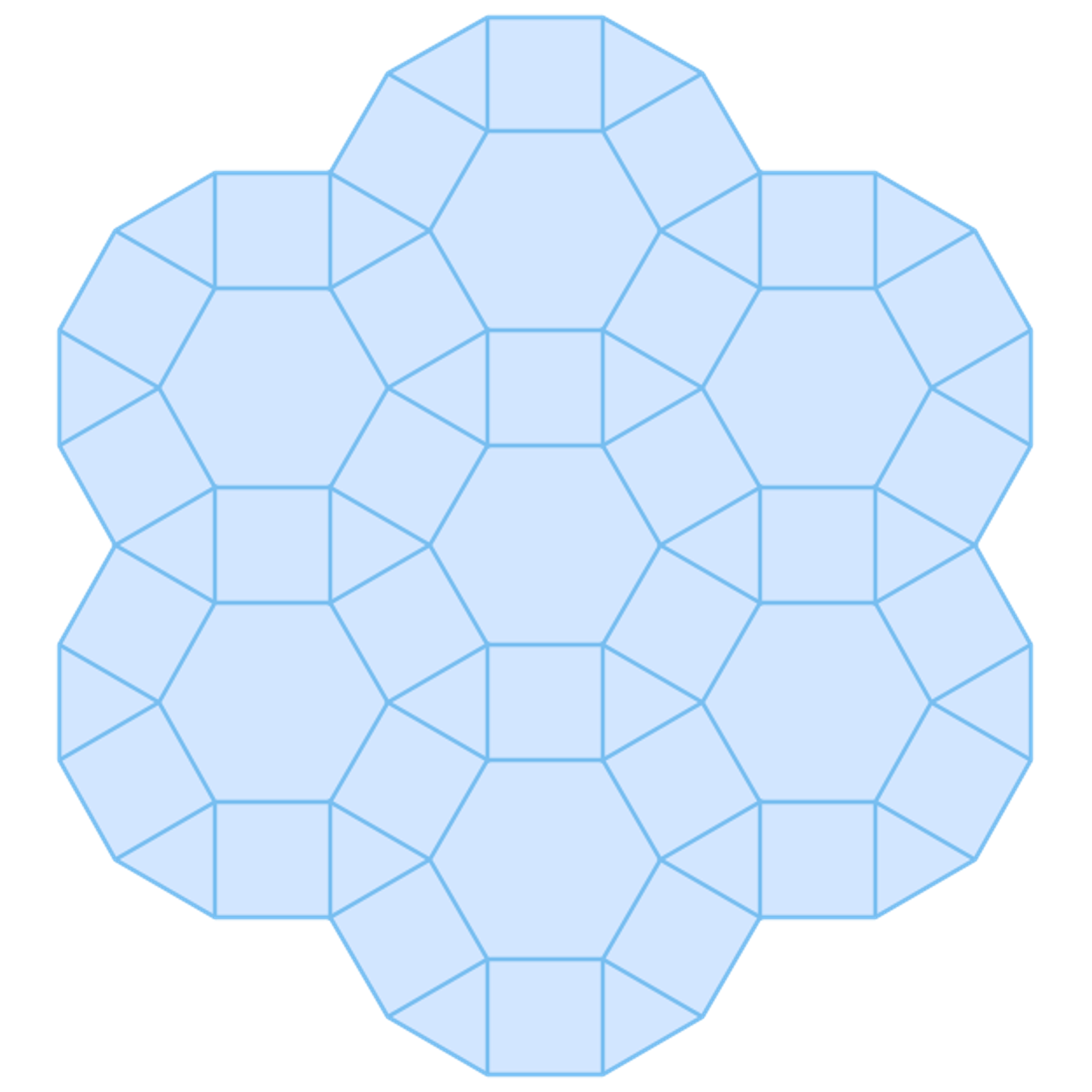}
    }
% 	\includegraphics[width=\textwidth]{figs/tiling-3464.png}
%         	\subcaption{} 
%         	\label{fig:SubdivideA}
\end{minipage}
%\hspace{.1cm}
\begin{minipage}[b]{0.32\textwidth} 
    \subfloat[\label{fig:SubdivideB}]{%
        \includegraphics[width=.95\textwidth]{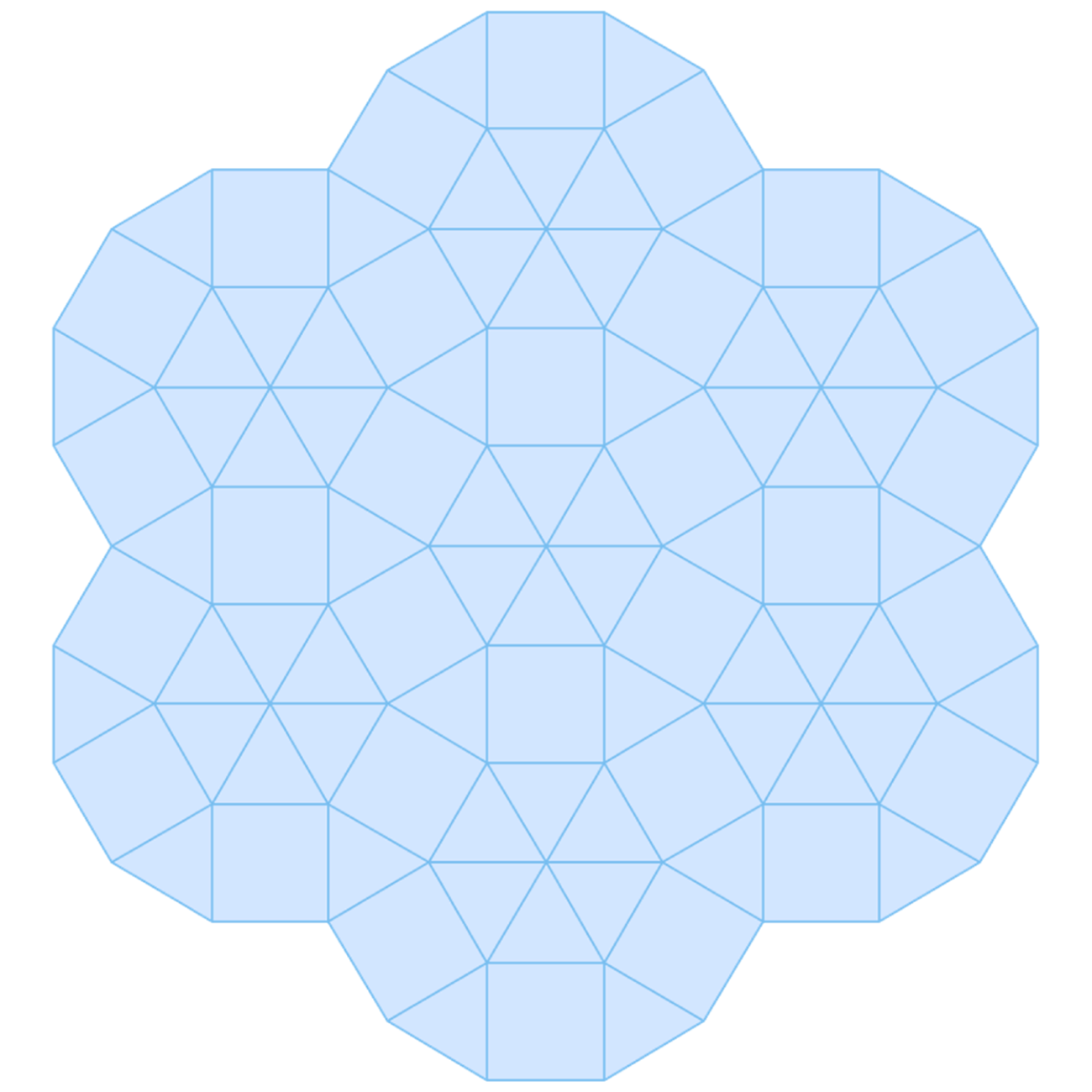}
    }
% 	\includegraphics[width=\textwidth]{figs/exotic-1.png}
%         	\subcaption{} 
%         	\label{fig:SubdivideB}
\end{minipage}
%\hspace{.1cm}
\begin{minipage}[b]{0.32\textwidth} 
    \subfloat[\label{fig:SubdivideC}]{%
        \includegraphics[width=.95\textwidth]{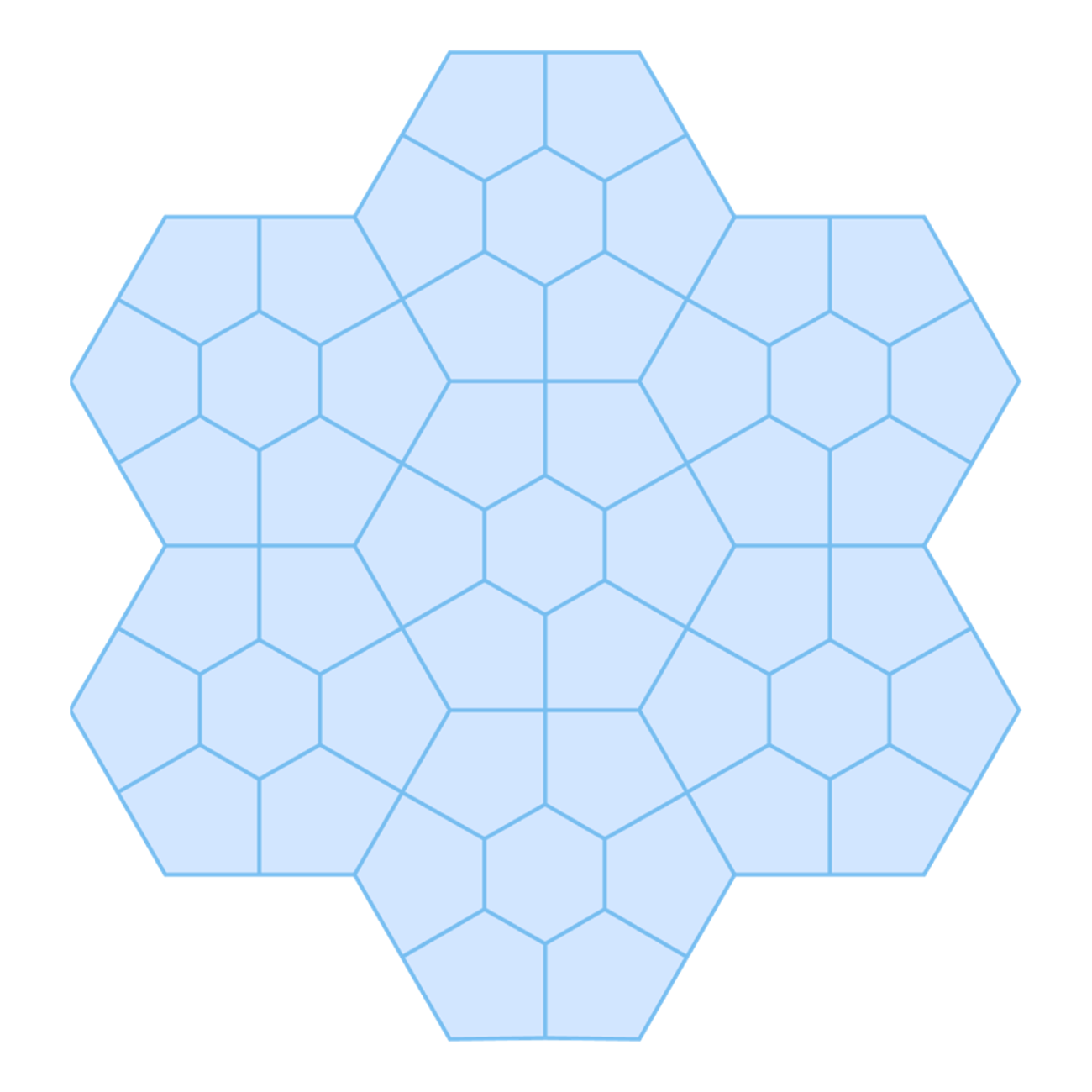}
    }
% 	\includegraphics[width=\textwidth]{figs/exotic-2.png}
%         	\subcaption{} 
%         	\label{fig:SubdivideC}
\end{minipage}
\hspace{0.5cm}
\begin{minipage}[b]{0.49\textwidth} 
    \subfloat[\label{fig:SubdivideD}]{%
        \includegraphics[width=1\textwidth]{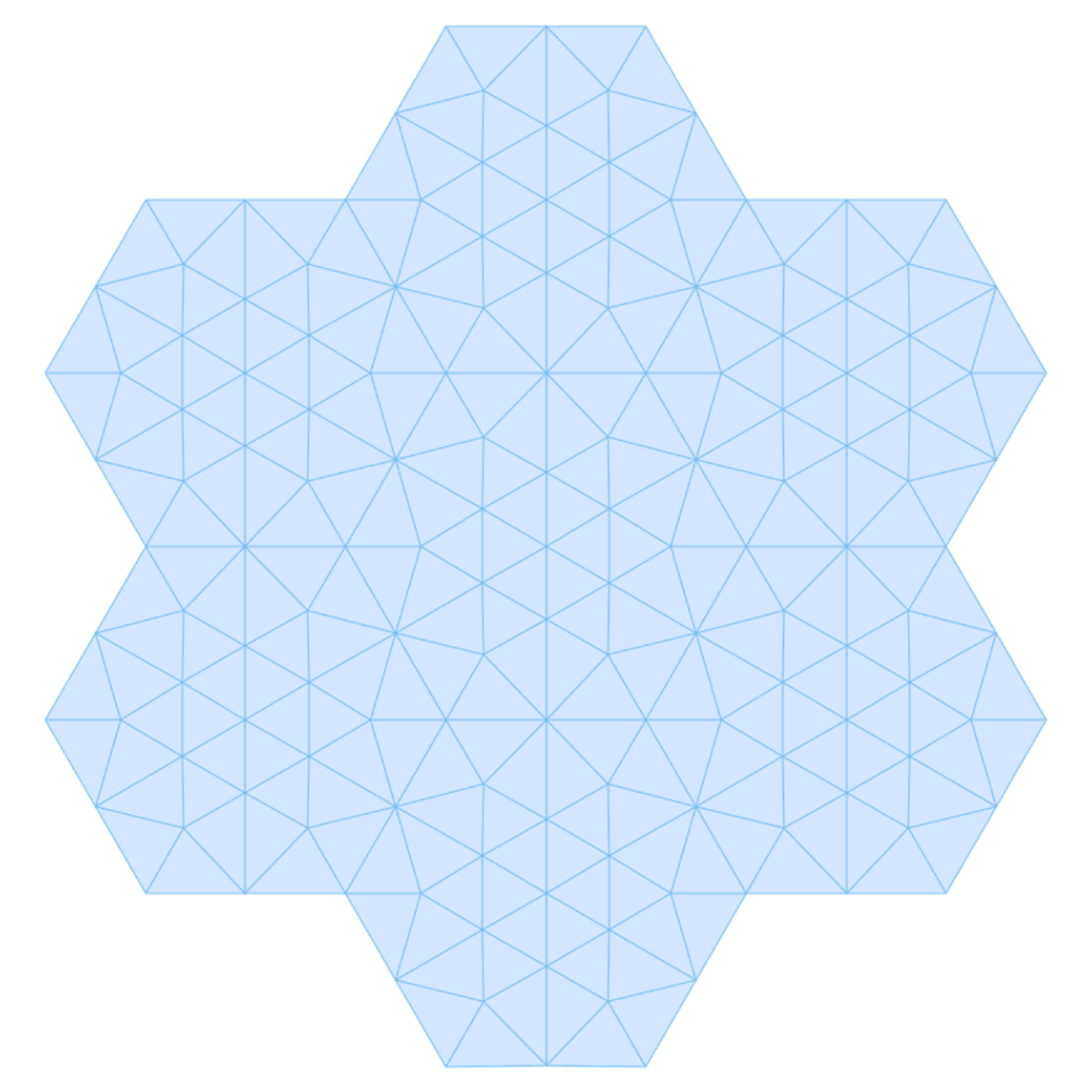}
    }
% 	\includegraphics[width=\textwidth]{figs/exotic-3.png}
%         	\subcaption{} 
%         	\label{fig:SubdivideD}
\end{minipage}
%\hspace{0.5cm}
\begin{minipage}[b]{0.49\textwidth} 
    \subfloat[\label{fig:SubdivideE}]{%
        \includegraphics[width=1\textwidth]{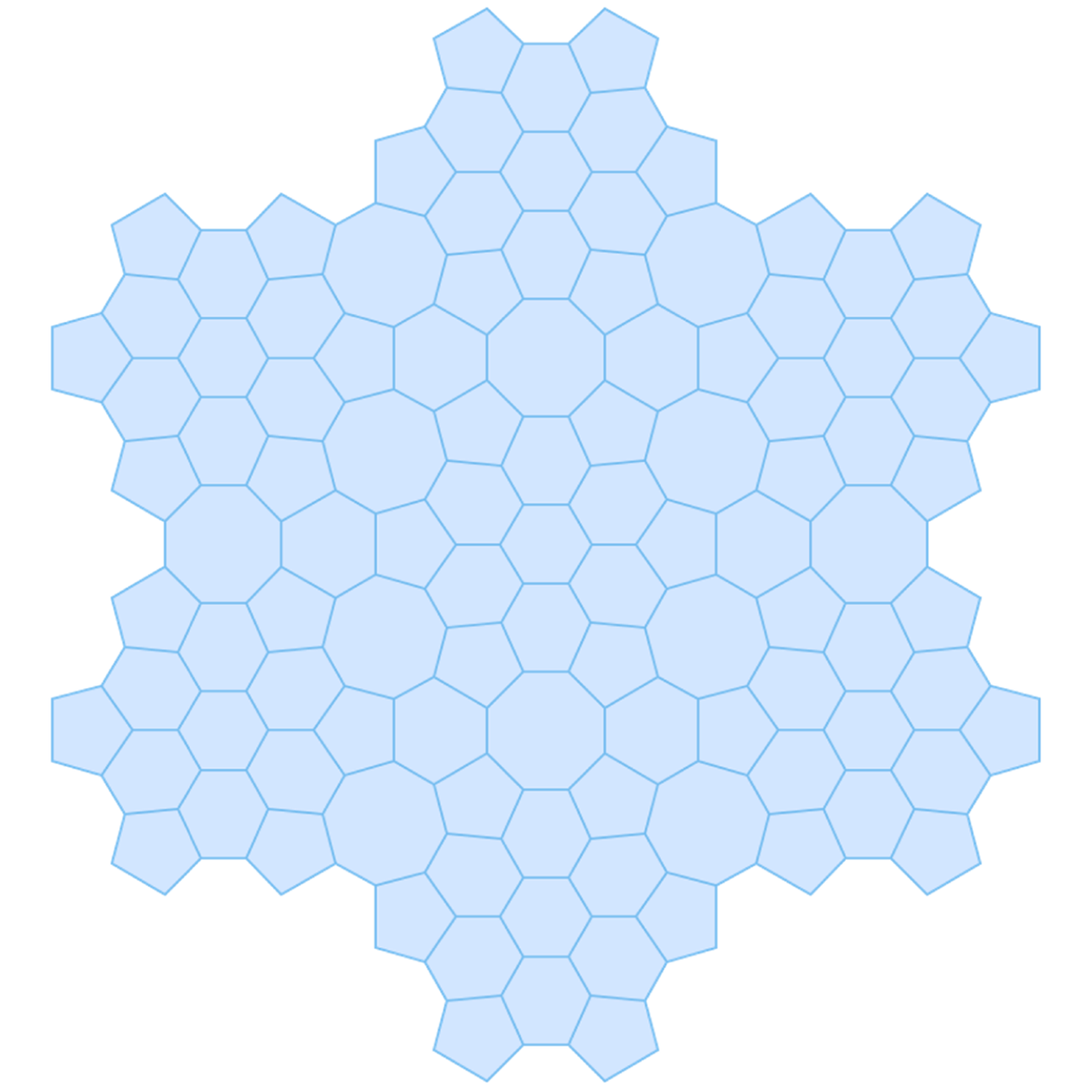}
    }
% 	\includegraphics[width=\textwidth]{figs/exotic-4.png}
%         	\subcaption{} 
%         	\label{fig:SubdivideE}
\end{minipage}
\caption{A 3.4.6.4 tiling (a) subdivided at $N \geq 6$ (b), its dual (c), all subdivided (d) and its dual (e).}
\label{fig:Subdivide}
\end{figure}

\noindent
Another useful operator is {\tt subdivide}, which subdivides all faces with $N$ or more sides into triangular sub-faces that share a central vertex (default $N = 1$). 
Figure \ref{fig:Subdivide} shows a sequence of {\tt subdivide} and {\tt dual} operations applied to a rhombitrihexahedral 3.4.6.4 tiling to produce a novel and exotic board design:

\phantom{}

{\tt (dual (subdivide (dual (subdivide (tiling T3464 2) min:6)))) }

\section{Conclusion}

\noindent
The Ludii grammar provides a simple way to describe most conceivable game boards by their underlying graphs, using tiling, shape and graph operators. 
This approach has allowed us to model the boards of over a thousand games for the Ludii general game system, and continues to produce interesting new board designs based on simple operations.

Future work will include improvements to symmetric board colourings (for games in which cell colour is relevant) and adaptive coordinate labelling that follows the contours of exotic boards. 
But the inclusion of a freeform {\tt graph} ludeme means that almost any game board that can described as a combination of vertices, edges and/or cells can be defined, making this approach ideal for the wide range of game boards to be modelled for the Digital Ludeme Project.

%%%%%%%%%%%%%%%%%%%%%%%%%%%%%%%%%%%%%%%%%%%

\section*{Acknowledgements}

This research was funded by the European Research Council as part of the Digital Ludeme Project (ERC CoG \#771292) led by Cameron Browne at Maastricht University's Department of Data Science and Knowledge Engineering. %over 2018–2023.

%%%%%%%%%%%%%%%%%%%%%%%%%%%%%%%%%%%%%%%%%%%

\bibliographystyle{splncs04}
\bibliography{dlp-biblio-1}

\end{document}